\documentclass{article}

\usepackage{graphicx}

\usepackage{caption}
\usepackage{amsmath}
\usepackage{amsfonts}
\usepackage{stmaryrd}
\usepackage{proof}
\usepackage{array}
\usepackage{tikz}
\usetikzlibrary{matrix,arrows,shapes,positioning}
\usepackage{lscape}

\newcommand{\references}{jlm-bibdata}

\newcommand{\nc}{\newcommand}
\nc{\nldia}{\Diamond}
\nc{\nlbox}{\Box}
\nc{\fdia}{\Diamond}
\nc{\gbox}{\Box}
\nc{\bs}{\backslash}
\nc{\pijl}{\rightarrow}
\nc{\arr}[2]{#1\pijl#2}
\nc{\dpijl}{\longleftrightarrow}
\nc{\rarepijl}{\rightsquigarrow}
\nc{\comp}{\circ}
\nc{\vektor}[1]{\overrightarrow{#1}}
\nc{\bra}{\langle}
\nc{\ket}{\rangle}
\nc{\braket}[2]{\bra #1\ |\ #2 \ket}

\nc{\cat}{\mathcal{C}}
\nc{\cattwo}{\mathcal{D}}
\nc{\focus}[1]{\fbox{$#1$}}
\nc{\seq}[2]{#1\Rightarrow #2}
\nc{\stripe}{\ | \ }
\nc{\app}[2]{(#1 \ #2)} 
\nc{\abs}[2]{\lambda #1.#2}
\nc{\blue}[1]{\textcolor{blue}{#1}}
\nc{\red}[1]{\textcolor{red}{#1}}
\nc{\s}{\slash}
\nc{\tensor}{\otimes}
\nc{\vecspace}[1]{\mathbf{#1}}
\nc{\resright}{\rhd}
\nc{\resleft}{\lhd}
\nc{\resrightinv}{\resright^{\minus1}}
\nc{\resleftinv}{\resleft^{\minus1}}
\nc{\meaninginterpretationF}[1]{\left[ #1 \right]}
\nc{\meaninginterpretation}[1]{\llbracket #1 \rrbracket}
\nc{\quotes}[1]{``#1''}
\nc{\reals}{\mathbb{R}}
\nc{\powersetvectorspace}{V_{P(U)}}
\nc{\means}[1]{\meaninginterpretation{\text{#1}}}
\newcommand{\plus}{\raisebox{.2\height}{\scalebox{.8}{+}}}
\newcommand{\minus}{\raisebox{.0\height}{\scalebox{1}{-}}}
\newtheorem{definition}{Definition}

\begin{document}

\title{A Proof-Theoretic Approach to Scope Ambiguity in Compositional Vector Space Models\footnote{This is a preprint of a paper to appear in: Journal of Language Modelling, 2018.}}


\author{Gijs Jasper Wijnholds}


\date{School of Electronic Engineering and Computer Science,\\ Queen Mary University of London \\ \today}


\maketitle

\begin{abstract}
We investigate the extent to which compositional vector space models can be used to account for scope ambiguity in quantified sentences (of the form \emph{Every man loves some woman}). Such sentences containing two quantifiers introduce two readings, a direct scope reading and an inverse scope reading. This ambiguity has been treated in a vector space model using bialgebras by \cite{hedges2016generalised} and \cite{sadrzadeh2016quantifier}, though without an explanation of the mechanism by which the ambiguity arises. We combine a polarised focussed sequent calculus for the non-associative Lambek calculus \textbf{NL}, as described in \cite{moortgat2011proof}, with the vector based approach to quantifier scope ambiguity. In particular, we establish a procedure for obtaining a vector space model for quantifier scope ambiguity in a derivational way.
\end{abstract}

\newpage

\section{Introduction}

There is a long standing tradition in formal semantics on compositionality: to separate the meaning of basic elements (lexical semantics) from the construction of higher-level meaning (derivational semantics) one assigns a homomorphism from a \emph{syntactic algebra} to a \emph{semantic algebra}. Having been rigorously formalised by Montague in his seminal papers \cite{montague1970english,montague1973proper}, these ideas have been made concrete in the field of typelogical grammar, where syntactic types are mapped onto semantic types so that any derivation gives rise to a \emph{meaning recipe}. Traditionally, meaning is taken to be a linear lambda term that evaluates to a truth value.

Ongoing research on distributional semantics, based on the idea that word meaning is defined relative to a word's context, has revealed an appealing way to incorporate typelogical grammar into distributional models \cite{coecke2010mathematical}. This approach, also known as the DisCoCat approach (Distributional Compositional Categorical models) treats compositionality in the Montagovian style as a functorial passage from syntactic types and proofs to vectors and linear maps. Given that this line of research is still in its early phase, there is much to be done to formalise details of the model, give accounts for semantic phenomena, and evaluate the effectiveness of the chosen approach. 

Though traditional categorial syntax and semantics go hand in hand, some aspects of the set-theoretic formal semantics go lost in the switch to a vector space model of meaning. First, the interpretation of constants that one can appeal to in a formal semantics are not directly available in a vector based setting; a logical word like \quotes{not} can be computed in the formal setting by taking set complement, but negating a vector or matrix is not trivial\footnote{Though there is work on simulating negation in a tensor based setting \cite{grefenstette2013towards}, it is not clear what negation really means in a distributional setting. For instance, an alternative view is to treat distributional negation as conversational \cite{kruszewski2016there}.}. Similarly, for coordinators like \quotes{and} and \quotes{or} the standard set intersection and union are not available in a vectorial setting. One could replace intersection by vector multiplication and union by vector summation, but in the presence of concrete distributional vectors it is not clear that such operations indeed perform well in an experimental setting. Second, the DisCoCat approach assumes a tight categorical correspondence between a syntactic formalism and the concrete vector semantics: when we want to stay in the realm of finite dimensional vector spaces, we are dealing with a compact closed category; to model a categorial grammar as a category, one needs to fully explicate its proof-theoretic logical and structural rules, an exposition that is not trivially available for any categorial system\footnote{For instance, the composition and type-raising combinators one finds in Combinatorial Categorial Grammar \cite{steedman2000syntactic} don't easily translate into a standard category, and the Displacement Calculus of \cite{morrill2011displacement} subsumes it's structural rules in the rules of the system \cite{valentin2014hidden}.}. Another issue with this categorical treatment is that a simple vector based model does not have the non-linearity that some models would assume. As an example, allowing a non-linearity in lexical lambda terms or as a syntactic mechanism means the copying of material which is not possible with all vectors. We discuss this issue in more detail in the rest of the paper. 

Some of the above issues have been addressed in recent work by \cite{sadrzadeh2013frobenius,hedges2016generalised,sadrzadeh2016quantifier}, giving accounts of subject/object relativisation, generalised quantifiers, and quantifier scope.
In \cite{sadrzadeh2013frobenius}, the meaning of pronoun relative clauses is explained by using Frobenius algebras in the lexicon, and assigning different pregroup grammar types to the subject relative pronoun \quotes{who} and the object relative pronoun \quotes{whom}. Two different derivations then naturally arise, giving an intersectional meaning to subject relative clauses like \quotes{Men who like Mary}, and object relative clauses like \quotes{Men whom Mary likes}. Such an approach does not lend itself to certain Germanic languages where the ambiguity has to be derivational: in Dutch, both the subject relative and object relative interpretations share the surface form \quotes{Mannen die Marie mogen}. To deal with this without specifying lexical alternatives, i.e. different possible typings of the relative pronoun \quotes{die}, \cite{moortgat2017lexical} provide a derivational account that results in the same intersective vector space meaning as the ones of \cite{sadrzadeh2013frobenius}.

An element that lacks in the results obtained so far on quantifier scope ambiguity is a detailed discussion of the derivational process, giving rise to ambiguities. Quantifier scope ambiguity as opposed to pronoun relativisation is more pressing as the scope ambiguity exists in English, and does not come from the lexicon, but rather different ways of reading the same surface form. The account of \cite{hedges2016generalised} explores the use of bialgebras to represent quantifiers, using context free grammars as the syntactic engine; its follow up discusses scope ambiguity but assumes the ambiguity to be given before detailing the direct scope and inverse scope readings of phrases of the shape \quotes{Every man loves some woman}. In order to explain how the ambiguity comes about, we need to detail the syntactic process, and integrate it with a vector based semantics. 

 Our goal in this paper, then, is to pave the way to fully explain compositionality in vector space models of meaning while also taking into account the desirable mechanisms of e.g. Frobenius algebras and bialgebras. Our step in this paper, is to show how we can represent quantifier scope ambiguity in a derivational manner, fully determined by the syntactic process combined with a suitable lexical semantics.
 
We will make use of a polarised non-associative Lambek calculus, and use focussing as a technique to gain control over the space of sequent derivations. A \emph{continuation-passing-style} translation from syntactic types into semantic objects then gives rise to the expected reading for quantifier scope ambiguity. This technique has been worked out by \cite{moortgat2011proof} (following \cite{bernardi2010continuation} and \cite{bastenhof2012polarized}), but has not until now been put in the context of vector space models.

This paper is structured as follows: in Section 2, we will briefly discuss quantifier scope ambiguity and its apparent non-linearity. Next, in Section 3 we define the basic, compositional DisCoCat model. We proceed to review quantifier scope ambiguity in vector space models in Section 4, and show in Section 5 how we can derive quantifier scope ambiguity in a compositional way using a polarised focussed sequent calculus that is interpreted in a vector space model. We conclude in Section 6 by explaining how our results can be further expanded and we introduce some potential new areas of investigation.

\section{Quantifier Scope Ambiguity}

There seems to be an intrinsic non-linearity associated with quantifiers. Consider the word \quotes{all} in a phrase \quotes{all men sleep}. One way of modelling the universal quantification in the phrase is to let \quotes{all} refer to an operation that decides whether the set of \quotes{men} is a subset of those entities that are sleeping, i.e. if \quotes{men} refers to some set $A$, and \quotes{sleep} to some set $B$, then \quotes{all men sleep} computes whether $A \subseteq B$. This can be given an alternative definition:
	$$
	\meaninginterpretationF{all}(A)(B) =
	\begin{cases}
		1 & \text{if}\ A = A \cap B \\
		0 & \text{otherwise}
	\end{cases} 
	$$
When one tries to give this interpretation in terms of a $\lambda$-term, the usual approach is to model both \quotes{men} and \quotes{sleep} as a \emph{characteristic function} of a set of entities, where \quotes{all} will be given a non-linear $\lambda$-term:
$$ \meaninginterpretation{all} = \abs{P}{\abs{Q}{\app{\forall}{\abs{x}{\app{Q}{x} \pijl \app{P}{x}}}}} $$
This $\lambda$-term will effectively decide whether $A \subseteq B$, or alternatively whether $A = A \cap B$.
Both the modellings sacrifice linearity in a sense: where the first, relational interpretation needs to use $A$ as an operand to the intersection operation and as an argument to decide equality, the second interpretation has to \emph{copy} the variable $x$ to decide whether everything in the universe satisfying the property $Q$ also satisfies $P$. We argue that this required non-linearity that is introduced by allowing non-linear $\lambda$-terms to be inserted through the lexicon, is exactly the same kind of non-linearity that is introduced to vector space models by means of bialgebra operations.
	It has been argued before that modelling quantification in vector space models forces one to use non-linear maps \cite{grefenstette2013towards}. However, this issue has been partially resolved by \cite{hedges2016generalised} when one admits a powerset structure to the basis vectors of the model. The then obtained bialgebra operations are linear in the algebraic sense, but non-linear in terms of typing information. That is, they allow for copying a resource $X$ into a resource $X \tensor X$ and deleting a resource in the opposite direction.
	That this kind of operation would jeopardize a Lambek style grammar formalism is immediate as the bialgebra operations would correspond to contraction and expansion, respectively.
	Our argument will proceed by claiming that a continuation-passing-style translation that allows for lexical insertion of non-linear $\lambda$-terms can instead be interpreted by means of the bialgebra operations of \cite{hedges2016generalised}.

\section{Compositional Distributional Semantics}

Compositional distributional semantics in a categorical setting takes a mathematically rigorous approach to compositionality. Much like traditional Montagovian semantics, there is a syntactic algebra involved that provides grammaticality by means of a proof system, in this case it can be either \emph{pregroup grammar} or the \emph{Lambek calculus} \cite{lambek1958mathematics,lambek1997type}. The \emph{semantic algebra} is, in the basic setup, the category of finite dimensional vector spaces, denoted \textbf{FVect}: content words are assigned a vector that represents its position in the space of word meanings, obtained through some method of co-occurrence extraction on a corpus. Whenever a sequence of words, annotated with their \emph{syntactic types}, leads to a derivation that proves grammaticality, the proof term associated with that derivation provides a linear map on the vectors associated with basic words, which after evaluation gives us the \emph{phrase meaning} of that sequence of words.

\subsection{Lambek grammars}
\label{sec:lambekgrammars}

	We make the model sketched above concrete by giving the relevant definitions. These are based on work by \cite{wijnholds2014categorical} in combination with the work of \cite{coecke2013lambek}.
	
\begin{definition}[Lambek types] Given a set $T$ of basic types, the set of Lambek types $F(T)$ is the smallest set such that:
\begin{enumerate}
	\item If $p \in T$ then $p \in F(T)$,
	\item If $A,\ B \in F(T)$ then $A \tensor B, \ A \backslash B, \ B \slash A \in F(T)$.
\end{enumerate}
\end{definition}	

\noindent We proceed to define a Lambek calculus in terms of a labelled deductive system, i.e. we use the notation of an inference system to show how proofs are derived:

\begin{definition}[Non-associative Lambek calculus] The (non-associative, non-unitary) Lambek calculus \textbf{NL} over $T$ is given by the types in $F(T)$ and the proofs generated by the following (labelled) inference system:

\begin{center}
\begin{tabular}{cc}
	$\infer[Ax]{1_A : A \pijl A}{}$ & $\infer[T]{g \comp f : A \pijl C}{f : A \pijl B & g : B \pijl C}$ \\
	& \\
	$\infer[R1]{\resright f : A \pijl C \s B}{f : A \tensor B \pijl C}$ & $\infer[R2]{\resleft f : B \pijl A \bs C}{f : A \tensor B \pijl C}$ \\
	& \\
	$\infer[R1^{\minus1}]{\resright^{\minus1} g : A \tensor B \pijl C}{g : A \pijl C \s B}$ & $\infer[R2^{\minus1}]{\resleft^{\minus1} g : A \tensor B \pijl C}{g : B \pijl A \bs C}$ \\
\end{tabular}
\end{center}
\end{definition}

\noindent One can show that monotonicity laws for each of the connectives are derived rules of inference:

\begin{center}
	$\infer[M_\tensor]{f \tensor g : A \tensor B \pijl C \tensor D}{f : A \pijl C & g : B \pijl D}$ \\
	\ \\
	$\infer[M_\s]{g \s f : B \s C \pijl D \s A}{f : A \pijl C & g: B \pijl D}$ \\
	\ \\
	$\infer[M_\bs]{f \bs g : C \bs B \pijl A \bs D}{f : A \pijl C & g : B \pijl D}$
\end{center}
where we have that
\begin{center}
	$f \tensor g := \resrightinv (( \resright \resleftinv (( \resleft 1_{C \tensor D} ) \comp g )) \comp f )$\\
	$g \s f := \resright (g \comp ( \resleftinv (( \resleft \resrightinv 1_{B \bs C} ) \comp f)))$ \\
	$f \bs g := \resleft (g \comp ( \resrightinv (( \resright \resleftinv 1_{C \bs B}) \comp f )))$
\end{center}

\noindent Leaving aside the issue of global associativity and its desirability from a linguistic perspective, we note how it can be added using two additional axioms:

\begin{center}
		\begin{tabular}{c}
			$\infer[Ass]{a_{A,B,C} : (A \tensor B) \tensor C \pijl A \tensor (B \tensor C)}{}$ \\
			\ \\
			$\infer[Ass^{\minus1}]{a^{\minus1}_{A,B,C} : A \tensor (B \tensor C) \pijl (A \tensor B) \tensor C}{}$
		\end{tabular}
\end{center}

\noindent The \emph{categorical} version of the Lambek calculus can be obtained by imposing the relevant standard equivalences on proofs, amongst others stipulating that composing with the identity proof is a vacuous operation, and that all two-way inference rules are isomorphims. For more detail we refer the reader to \cite{wijnholds2014categorical}.

To make grammaticality judgments to sequences of words, we need a \emph{lexicon} assigning types to words over an alphabet. For the sake of completeness we define the lexicon as a relation, but in the remainder of this paper we will freely abuse notation and treat the lexicon as if it were a function.

\begin{definition}[Lexicon] Let $\Sigma$ be a finite, non-empty set of words (an alphabet). A lexicon over $\Sigma$ is a relation $\delta \subseteq \Sigma \times F(T)$.
\end{definition}

\begin{definition}[Lambek grammar] Given a set of basic types $T$, a Lambek grammar over $T$ is  a triple $(\Sigma, \delta, S)$ where $\Sigma$ is an alphabet, $\delta$ is a lexicon over $T$, and $S \in F(T)$ is a distinguished goal type.
\end{definition}

\begin{definition}[Grammaticality] Given a Lambek grammar $(\Sigma, \delta, S)$ over $T$, we say that a sequence of words $w_1 ... w_n$ over $\Sigma$ is grammatical iff there is a merged sequence $W_1 \tensor W_2 ... \tensor W_n$ (where for each $i$ we have $w_i \delta W_i$), and there exists a proof of $W_1 \tensor W_2 ... \tensor W_n \pijl S$ in the Lambek calculus.
\label{def:grammaticality}
\end{definition}

\noindent The presented definitions so far give a procedure to obtain a proof of sentencehood for a sequence of words. Moreover, there might be several proofs of the same sequence of words. This may be desirable (in cases of derivational ambiguity) or not (in the case proof-theoretic redundancy, e.g. the successive to and fro use of two-way rules). In the categorical variant of the Lambek calculus, we can simply take the proofs of sentencehood of a sequence to be the hom-set of morphisms $Hom(W_1 \tensor W_2 ... \tensor W_n,S)$. This produces fewer proofs as unnecessary ambiguity of the proof system is brought down by categorical equations.
The structure of the (non-associative) Lambek calculus \textbf{NL} is that of a \emph{biclosed magmatic category}\footnote{A magmatic category is a weaker version of a monoidal category: the tensor has no unit and is not necessarily associative. See \cite{wijnholds2017coherent}.}.

\subsection{Finite dimensional vector space models}

Lambek grammars are easily interpretable in vector space semantics as vector spaces enjoy compact closure, a weaker variant of the bi-closure of the Lambek calculus. We define the category \textbf{FVect} and show that it enjoys compact closure:

\begin{definition}[Compact Closure] A  compact closed category is a monoidal category $\cat$ with dual objects $A^l,A^r$ for every object $A$ in $\cat$ and additional morphisms
	$$ A^l \tensor A \xrightarrow[]{\epsilon^l_A} I \xrightarrow[]{\eta^l_A} A \tensor A^l $$
	$$ A \tensor A^r \xrightarrow[]{\epsilon^r_A} I \xrightarrow[]{\eta^r_A} A^r \tensor A $$
that satisfy the yanking properties\footnote{Note that we left out the hidden associativity morphism.}

\begin{center}
\begin{tabular}{cc}
	$(id_A \tensor \epsilon^l_A) \comp (\eta^l_A \tensor id_A) = id_A$ & $(\epsilon^r_A \tensor id_A) \comp (id_A \tensor \eta^r_A) = id_A$ \\[0.75em]
	$(\eta^l_A \tensor id_{A^l}) \comp (id_{A^l} \tensor \eta^l_{A}) = id_{A^l}$ & $(id_{A^r} \tensor \eta^r_A) \comp (\eta^r_A \tensor id_{A^r}) = id_{A^r}$ \\
\end{tabular}
\end{center}
\end{definition}

\noindent In the category of finite dimensional vector spaces \textbf{FVect} we have that the dual space $A^*$ is isomorphic to $A$ when we fix a basis (which is the case for concrete models). The $\epsilon$ and $\eta$ maps, now reduced to just two maps, are given by

\begin{center}
\begin{tabular}{c}
	$\epsilon := \sum_{ij} c_{ij} (v_i \tensor v_j) \mapsto \sum_{ij} c_{ij} \braket{v_i}{v_j}$ \\
	\ \\
	$\eta := 1 \mapsto \sum_{i} (v_i \tensor v_i)$
\end{tabular}
\end{center}

\noindent In concrete vector models, we will have vectors learnt for content words. For instance, the noun phrases \quotes{John} and \quotes{Mary} can be interpreted as vectors $\vektor{n_1},\vektor{n_3} \in \mathbf{N}$, respectively. This means that they are essentially single points in a vector space. Setting the sentence space to be the real numbers, a transitive verb like \quotes{loves} would live in the vector space $\mathbf{N} \tensor \reals \tensor \mathbf{N}$, and would carry information about the degree with which individuals love one another. In vector terms: $$\sum\limits_{ij} c_{ij} (\vektor{n_i} \tensor 1 \tensor \vektor{n_j})$$ The $c_{ij}$ is the respective degree for any pair of individuals $i,j$. The meaning of the phrase \quotes{John loves Mary} should then reduce to taking the inner product of the noun phrases with the verbs and so should give $$\sum\limits_{ij} c_{ij} \braket{\vektor{n_1}}{\vektor{n_i}}\braket{\vektor{n_j}}{\vektor{n_3}}$$ In the next section, we show how to relate derivations in a Lambek grammar to concrete computations in a vector space model.

\subsection{Interpretation}

Given that the compact closedness of \textbf{FVect} instantiates the closure of the Lambek calculus, we can easily interpret proofs in a Lambek grammar in a vector space model by passing from words and their lexical types to vectors in a homomorphically obtained vector space. Any proof of grammaticality will be interpreted through the $\eta$ and $\epsilon$ maps: \begin{definition}[Interpretation] Let $(\Sigma, \delta, S)$ be a Lambek grammar over $T$. An \emph{interpretation} is a pair of maps $I_0 : F(T) \pijl \mathbf{FVect}, I_1 : \Sigma \pijl \delta(\Sigma)$, where $\delta(\Sigma)$ is the relational image of $\delta$, such that $I_0$ respects typing and $I_1$ respects lexical type assignment. That is,
\begin{center}
\begin{tabular}{c}
$ I_0(A \tensor B) = I_0(A \bs B) = I_0(A \s B) = I_0(A) \tensor I_0(B) $ \\[0.75em]
and \\[0.75em]
$ I_1(w) = \vektor{v} \quad \text{ iff } \quad w \delta W \text{ and } \vektor{v} \in I_0(W)$
\end{tabular}
\end{center}
\end{definition}

\noindent An interpretation map sends words to vectors that respect the syntactic types associated with those words. We need to give a vectorial interpretation of proofs as well, in order to know how to compute meanings of a tuple of vectors.
The identity proof and transitivity of proofs carries over to the identity map on vector spaces and the composition of linear maps. The remaining rules of residuation are interpreted as shown below:

\begin{center}
\begin{tabular}{c}
	$ \infer[R1]{ \big(f' \tensor id_{I_0(B)}\big) \comp \big(id_{I_0(A)} \tensor \eta_{I_0(B)}\big) : I_0(A) \pijl I_0(C) \tensor I_0(B)}{f' : I_0(A) \tensor I_0(B) \pijl I_0(C)}$ \\
	\ \\	
	$\infer[R2]{\big(id_{I_0(A)} \tensor f'\big) \comp \big(\eta_{I_0(A)} \tensor id_{I_0(B)}\big) : I_0(B) \pijl I_0(A) \tensor I_0(C)}{f' : I_0(A) \tensor I_0(B) \pijl I_0(C)}$ \\
	\ \\
	$\infer[R1^{\minus1}]{\big(id_{I_0(C)} \tensor \epsilon_{I_0(B)}\big) \comp \big(g' \tensor id_{I_0(B)}\big) : I_0(A) \tensor I_0(B) \pijl I_0(C)}{g' : I_0(A) \pijl I_0(C) \tensor I_0(B)}$ \\
	\ \\
	$\infer[R2^{\minus1}]{\big(\epsilon_{I_0(A)} \tensor id_{I_0(C)}\big) \comp \big(id_{I_0(A)} \tensor g'\big) : I_0(A) \tensor I_0(B) \pijl I_0(C)}{g' : I_0(B) \pijl I_0(A) \tensor I_0(C)}$ \\
\end{tabular}
\end{center}
\ \\
\noindent It is a nice puzzle for the reader to verify that by the yanking equations, we preserve isomorphicity of residuation, for example one can show that the interpretation of $\resrightinv \resright f$ is equal to the interpretation of $f$.

\subsection{Illustration}

Recall that we have vectors for \quotes{John}, \quotes{Mary} and \quotes{loves} and we have an intended meaning of the phrase \quotes{John loves Mary}. We take a Lambek grammar over the set of basic types $\{np, s\}$, where $np$ will be interpreted as $\mathbf{N}$ and $s$ will be mapped to $\reals$. We define a lexicon as follows: 
\begin{center}
\begin{tabular}{c|c|c|c}
	$w$ & $\delta(w)$ & $I_1(w)$ & $I_0(\delta(w))$ \\
	\hline
	\quotes{John} & $np$ & $n_1$ & $\vecspace{N}$ \\ 
	\quotes{Mary} & $np$ & $n_3$ & $\vecspace{N}$ \\
	\quotes{loves} & $(np \bs s) \s np$ & $\sum_{ij} c_{ij} (\vektor{n_i}\tensor 1 \tensor \vektor{n_j})$ & $\vecspace{N \tensor \mathbb{R} \tensor N}$ \\
\end{tabular}
\end{center}

\noindent Given that $\resleftinv \resrightinv (1_{(np \bs s) \s np})$ proves grammaticality of \quotes{John loves Mary}, the associated meaning computation will be \\
\ \\
\begin{tabular}{l}
$ (\epsilon_N \tensor id_{\reals}) \comp (id_N \tensor ((id_{N \tensor \reals} \tensor \epsilon_N) \comp (id_{N \tensor \reals \tensor N} \tensor id_N)))$ \\[0.2em]
$\ \ \ \ (\vektor{n_1} \tensor  \sum_{ij} c_{ij} (\vektor{n_i}\tensor 1 \tensor \vektor{n_j}) \tensor \vektor{n_3})$ \\
\ \\
$ = (\epsilon_N \tensor id_{\reals}) \comp (id_{N \tensor N \tensor \reals} \tensor \epsilon_N) (\vektor{n_1} \tensor  \sum_{ij} c_{ij} (\vektor{n_i}\tensor 1 \tensor \vektor{n_j}) \tensor \vektor{n_3})$ \\
\ \\
$ = (\epsilon_N \tensor id_{\reals}) (\vektor{n_1} \tensor \sum_{ij} c_{ij} (\vektor{n_i}\tensor \braket{\vektor{n_j}}{\vektor{n_3}}) $ \\
\ \\
$ = \sum_{ij} c_{ij} \braket{\vektor{n_1}}{\vektor{n_i}}\braket{\vektor{n_j}}{\vektor{n_3}}  $ \\
\ \\
$ = c_{13} $ \\
\end{tabular}
\ \\
\ \\
\ \\
\noindent This result is exactly the intended meaning we wanted to obtain. Note that the result of the computation relies on the fact that the content words in the vector space model are taken to be the basis vectors, hence they are orthogonal. The result $c_{13}$ indicates the distributional strength of John loving Mary in a corpus that the vectors have been learnt from.
Until now, we have neglected discussion about function words: logical words, relative pronouns, and quantifiers are not intuitively represented well by co-occurrence data. The logical word \quotes{and} may occur with many different words, but that statistic does not tell us much about the meaning of the word. So although all the basic operations from a Lambek grammar are directly interpretable in vector space models, more advanced semantic phenomena lack an explanation in the simple models.

\section{Quantifier Scope Ambiguity in Vector Space Models}

In this section we review the use of bialgebras in vector space models as exhibited by \cite{hedges2016generalised,sadrzadeh2016quantifier} and show how the two scope readings can be obtained.
The treatment of quantifiers in vector space models relies on the use of powersets to function. As long as we can know of our vector space that its basis vectors are given by the powerset of some set $A$, we can perform additional operations on the vector space.

\begin{definition}[Bialgebra] Given a symmetric monoidal category $(\cat, \tensor, I, \sigma)$, a bialgebra on an object $X$ in $\cat$ is a tuple of maps
	$$ X \xrightarrow[]{\delta_X} X \tensor X \xrightarrow[]{\mu_X} X $$
	$$ X \xrightarrow[]{\iota_X} I \xrightarrow[]{\zeta_X} X $$
that satisfy the conditions of a monoid for $(X,\mu,\zeta)$ and a comonoid for $(X,\delta,\iota)$ and furthermore satisfy the bialgebra axioms:

\setlength\tabcolsep{3 pt}
\begin{center}
\begin{tabular}{lcl}
	$\iota \comp \mu$ & $=$ & $\iota \tensor \iota$ \\
	$\delta \comp \zeta$ & $=$ & $\zeta \tensor \zeta$ \\
	$\iota \comp \zeta$ & $=$ & $id_I$ \\
	$\delta \comp \mu$ & $=$ & $(\mu \tensor \mu) \comp (id_X \tensor \sigma \tensor id_X) \comp (\delta \tensor \delta)$ \\
\end{tabular}
\end{center}
\end{definition}
\setlength\tabcolsep{6 pt}
\noindent The last of the four equations tells us that in a bialgebra, the order of copying and merging is irrelevant given that we can switch copies by means of the symmetry of the category. What is interesting to note is that any powerset $P(U)$ bears a bialgebra structure if we consider the Cartesian product to be the tensor and the singleton set $\{ \star \}$ as the identity object. What follows is that any vector space over a powerset, denoted $V_{P(U)}$, carries a bialgebra structure. Both bialgebras are given below:

\begin{center}
\begin{tabular}{rcl@{\hskip 2cm}rcl}
	$A$ & $\overset{\delta}{\Leftrightarrow}$ & $A \times A$ & $|A\ket$ & $\overset{\delta}{\mapsto}$ & $|A\ket \tensor |A\ket$ \\
	$A \times B$ & $\overset{\mu}{\Leftrightarrow}$ & $A \cap B$ &	$|A\ket \tensor |B\ket$ & $\overset{\mu}{\mapsto}$ & $|A \cap B\ket$ \\
	$A$ & $\overset{\iota}{\Leftrightarrow}$ & $\{ \star \}$ &	$|A\ket$ & $\overset{\iota}{\mapsto}$ & $1$ \\
	$\{ \star \}$ & $\overset{\zeta}{\Leftrightarrow}$ & $U$ &	$1$ & $\overset{\zeta}{\mapsto}$ & $|U\ket$
\end{tabular}
\end{center}
The existence of a bialgebra on powerset vector spaces allows for a neat treatment of quantification. Given that nouns and noun phrases are represented as vectors on a powerset, universal quantification and existential quantification are treated as

\begin{center}
\begin{tabular}{rcl}
	$|A\ket$ & $\overset{\meaninginterpretation{all}}{\mapsto}$ & $\sum\limits_{A \subseteq B \subseteq U} |B\ket$ \\
	& & \\
	$|A\ket$ & $\overset{\meaninginterpretation{some}}{\mapsto}$ & $\sum\limits_{\substack{B \ \text{s.t.} \\ A \cap B \neq \emptyset}} |B\ket$
\end{tabular}
\end{center}
To get a feel for how the meaning of a quantified sentence should be computed according to \cite{hedges2016generalised}, we show the example of \quotes{all men sleep}, which gets assigned the meaning\\
\begin{center}
\begin{tabular}{l}
$ \epsilon_{\powersetvectorspace} \comp (\meaninginterpretation{\text{all}} \tensor \mu_{\powersetvectorspace}) \comp (\delta_{\powersetvectorspace} \tensor id_{\powersetvectorspace}) (|\meaninginterpretation{\text{men}}\ket \tensor |\meaninginterpretation{\text{sleep}}\ket)$ \\
\ \\
$ = \epsilon_{\powersetvectorspace} \comp (\meaninginterpretation{\text{all}} \tensor \mu_{\powersetvectorspace}) (|\meaninginterpretation{\text{men}}\ket \tensor |\meaninginterpretation{\text{men}}\ket \tensor |\meaninginterpretation{\text{sleep}}\ket)$ \\
\ \\
$ = \epsilon_{\powersetvectorspace} ( \sum\limits_{\meaninginterpretation{\text{men}} \subseteq B \subseteq U} |B\ket \tensor |\meaninginterpretation{\text{men}} \cap \meaninginterpretation{\text{sleep}}\ket)$ \\
\ \\
$ = \sum\limits_{\meaninginterpretation{\text{men}} \subseteq B \subseteq U} \bra B |\meaninginterpretation{\text{men}} \cap \meaninginterpretation{\text{sleep}}\ket $ \\
\ \\
$ = \bra \meaninginterpretation{\text{men}} |\meaninginterpretation{\text{men}} \cap \meaninginterpretation{\text{sleep}}\ket $ \\
\end{tabular}
\end{center}
\ \\
\noindent Although this approach works for statements with a single quantifier, it fails to deliver both reading for a doubly quantified statement such as \quotes{every student likes some teacher} as the computations for the subject and object quantifiers will be independent of each other. Hence, both readings will collapse to the same meaning.
	This lack of explanatory power of the model is amended in a subsequent paper \cite{sadrzadeh2016quantifier}, where the implicit quantified variable is passed on to the computation of the second quantifier. For a transitive verb such as \quotes{likes}, that is modelled as an element $$\meaninginterpretation{\text{likes}} = \sum\limits_{ij} c_{ij} (| A_i\ket \tensor | A_j\ket)$$ in $\powersetvectorspace \tensor \powersetvectorspace$, we can model the forward image of an element in $U$ as $$\meaninginterpretation{\text{likes}_a} = \sum\limits_{ij} w_{ij} \braket{\{a\}}{A_i} |A_j\ket$$ The backward image is computed similarly by taking the inner product of $\vektor{v_a}$ with $\vektor{v_j}$.
	This construction now allows for both readings of \quotes{every student likes some teacher}, though there is no procedure given to obtain these readings through a syntactic process.

\section{Quantifier Scope Ambiguity using Focussing and Polarisation}

Focussing is a proof-theoretic technique stemming from the work of \cite{andreoli2001focussing} that aims to eliminate redundancy from regular sequent systems. Focussed proof search proceeds by distinguishing those formulas that enjoy invertible introduction rules (\emph{asynchronous formulas}), and those that do not (\emph{synchronous formulas}). Asynchronous formulas are decomposed in a backward chaining proof search until there is no more decomposition possible. Then, one of the synchronous formulas is selected to be put in focus, after which the process of decomposition continues. This implies that now only the number of synchronous formulas determines the number of distinct proofs. This approach has been applied to the Lambek-Grishin calculus, a symmetric extension of the Lambek calculus, by \cite{bernardi2010continuation}, and is worked out in more in detail by \cite{moortgat2011proof}.

In order to obtain a compositional Montagovian semantics from a display style presentation of focussed proofs for the Lambek-Grishin calculus, \cite{bastenhof2012polarized} applies a polarisation technique, whereby formulas are assigned either positive or negative polarity. Atomic formulas are assigned an arbitrary polarity; the choice of this \emph{bias} affects the set of proofs obtained. The polarity also influences semantics: under the continuation semantics of \cite{bernardi2010continuation}, a negative formula will be \emph{negated} in its interpretation. Though the focussing and polarisation approach are described by \cite{bernardi2010continuation} and \cite{bastenhof2012polarized}, respectively, here we follow the focussed sequent presentation of \cite{moortgat2011proof}.

We start by defining polarity of types:
	
\begin{definition}[Polarity] Given a set of basic types $T$, a polarity assignment on types is a map $pol : F(T) \pijl \{ -,\plus \}$ that assigns to the types in $T$ an arbitrary polarity but fixes the polarity for complex types:

\begin{center}
\begin{tabular}{rcl}
	$pol(A \tensor B)$ & $=$ & $+$ \\
	$pol(A \bs B)$ & $=$ & $-$ \\
	$pol(B \s A)$ & $=$ & $-$ \\
\end{tabular}
\end{center}
\end{definition}

Given a Lambek grammar $G$ over a set $T$, grammaticality is defined similarly to Definition \ref{def:grammaticality}, where the set of proofs is given by the underlying proof system. The only difference is that the final sequent should have the consequent formula in focus. The proof is encoded by the abstract label of the proof, according to the abstract sequent system defined in Figure \ref{fig:seq}. 

\begin{figure}
\begin{center}
	\begin{tabular}{ll}
		\begin{tabular}{c}
		Focused types are positive \\
		\ \\
		\infer[Ax]{\blue{Ax(A,x)} \stripe \seq{x : A}{\focus{A}}}{} \\
		\ \\
		\ \\
		\infer[\leftharpoondown]{\blue{\leftharpoondown(M,x,A)} \stripe \seq{X[\focus{A}]}{Y}}{\blue{M} \stripe \seq{X[x : A]}{Y}} \\
		\ \\
		\ \\
		\infer[\rightharpoonup]{\blue{\rightharpoonup(M,\alpha)} \stripe \seq{X}{\alpha : A}}{\blue{M} \stripe \seq{X}{\focus{A}}} \\
		\end{tabular}
		&
		\begin{tabular}{c}	
		Focused types are negative \\	
		\ \\
		\infer[CoAx]{\blue{CoAx(A,\alpha)} \stripe \seq{\focus{A}}{\alpha : A}}{} \\
		\ \\
		\ \\
		\infer[\leftharpoonup]{\blue{\leftharpoonup(M,x)} \stripe \seq{X[x : A]}{Y}}{\blue{M} \stripe \seq{X[\focus{A}]}{Y}} \\
		\ \\
		\ \\
		\infer[\rightharpoondown]{\blue{\rightharpoondown(M,\alpha)} \stripe \seq{X}{\focus{A}}}{\blue{M} \stripe \seq{X}{\alpha : A}} \\
		\end{tabular}
	\end{tabular} \\
	\ \\
	\ \\
	\ \\
	\begin{tabular}{lr}
		\infer[\slash L]{\blue{\slash L(M,N)} \stripe \seq{X[\focus{A \slash B} \bullet Y]}{Z}}{\blue{M} \stripe \seq{X[\focus{A}]}{Z} & \blue{N} \stripe \seq{Y}{\focus B}}
		&
		\infer[\slash R]{\blue{\slash R(M,x,\alpha ,\beta)} \stripe \seq{X}{\beta : A \slash B}}{\blue{M} \stripe \seq{X \bullet x : B}{\alpha : A}} \\
		\ \\
		\infer[\otimes L]{\blue{\otimes L(M,x,y,z)} \stripe \seq{X[z : A \otimes B]}{Y}}{\blue{M} \stripe \seq{X[x : A \bullet y : B]}{Y}}
		&
		\infer[\otimes R]{\blue{\otimes R(M,N)} \stripe \seq{X \bullet Y}{\focus{A \otimes B}}}{\blue{M} \stripe \seq{X}{\focus{A}} & \blue{N} \stripe \seq{Y}{\focus{B}}} \\
		\ \\
		\infer[\bs L]{\blue{\bs L(M,N)} \stripe \seq{X[Y \bullet \focus{B \bs A}]}{Z}}{\blue{M} \stripe \seq{Y}{\focus{B}} & \blue{N} \stripe \seq{X[\focus{A}]}{Z}}
		&
		\infer[\bs R]{\blue{\bs R(M,x,\alpha ,\beta]} \stripe \seq{X}{\beta : B \bs A}}{\blue{M} \stripe \seq{x : B \bullet X}{\alpha : A}}
	\end{tabular}
\end{center}
\caption{Focussed labelled sequent system for \textbf{NL}}
\label{fig:seq}
\end{figure}

\subsection{CPS translation}

The translation of types and proofs given by \cite{moortgat2011proof} into a target semantic algebra is a two-step process:

\[\begin{array}{c}
\textsf{source}\\
\textbf{NL}_{\otimes, \bs, \slash}\\
\end{array} \stackrel{I}{\longrightarrow} 
\begin{array}{c}
\textsf{continuation}\\
\textsf{semantics}\\
\textbf{LP}_{\otimes,\bot}\\
\end{array}\stackrel{J}{\longrightarrow}
\begin{array}{c}
\textsf{target}\\
\textbf{FVect}\\
\end{array}
\]

Instead of considering a proof to be a simple transformation of values (the assumptions) to a value (the conclusion), we consider a proof to be a \emph{continuation}, a function that awaits an evaluation context to compute a final value. The intermediate semantics is the Lambek calculus with permutation and negation, \textbf{LP}$_{\otimes,\bot}$, a system that only uses a product operation but introduces a negation. Furthermore, permutation of resources is allowed to compensate for the lack of directionality without the $\s,\bs$ connectives. We will define a direct mapping from source to target, to skip the administrative details of the intermediate semantics.

In order to replicate the effect of the negation in \textbf{LP}$_{\otimes,\bot}$, we use vector spaces over sets; given some type $A$, we define its interpretation to be a vector space over a set. In this way, we enjoy the bialgebras defined over those vector spaces. First, a type $W$ is mapped to some set $A$, using the Cartesian product and powerset operations. Then, the final interpretation of a type will be the vector space over the given set, $V_A$. We get the intended tensor products on spaces due to the fact that $V_{A \times B} \cong V_A \tensor V_B$. 

\begin{definition}[Type interpretation] Given a set of basic types $T$ and a basic interpretation map $I_0 : T \pijl \mathbf{Set}$, the type interpretation is a map $I_1 : F(T) \pijl \mathbf{Set}$ defined as follows:

\begin{enumerate}
	\item For basic types $p \in T$ we have
		$$ I_1(p) = 
		\begin{cases}
			I_0(p) & \text{ if } pol(p) = + \\
			P(I_0(p)) & \text{ if } pol(p) = -
		\end{cases}
		$$
	\item For complex types, the interpretation depends both on the polarity of subtypes and the connective involved:
		\begin{center}
			\begin{tabular}{|cc | r@{\hskip -1em}c@{\hskip -1em}l | r@{\hskip -1em}c@{\hskip -1em}l | r@{\hskip -1em}c@{\hskip -1em}l |}
				\hline
				$A$ & $B$ & & $I_1(A \tensor B)$ & & & $I_1(A \bs B)$ & & & $I_1(B \s A)$ & \\
				\hline
				 $\minus$ & $\minus$ & $P(I_1(A))$ & $\times$ & $P(I_1(B))$ & $P(I_1(A))$ & $\times$ & $I_1(B)$ & $I_1(B)$ & $\times$ & $P(I_1(A))$ \\
				 $\minus$ & $\plus$ & $P(I_1(A))$ & $\times$ & $I_1(B)$ & $P(I_1(A))$ & $\times$ & $P(I_1(B))$ & $I_1(B)$ & $\times$ & $I_1(A)$ \\
				 $\plus$ & $\minus$ & $I_1(A)$ & $\times$ & $P(I_1(B))$ & $I_1(A)$ & $\times$ & $I_1(B)$ & $P(I_1(B))$ & $\times$ & $P(I_1(A))$ \\
				 $\plus$ & $\plus$ & $I_1(A)$ & $\times$ & $I_1(B)$ & $I_1(A)$ & $\times$ & $P(I_1(B))$ & $P(I_1(B))$ & $\times$ & $I_1(A)$ \\
				\hline
			\end{tabular}
		\end{center}
	\item We stipulate that for any type $A$, its interpretation $I_1(A)$ is lifted to the vector space spanned by its elements, that is we define the final interpretation $I_2 : F(T) \pijl \mathbf{FVect}$ as $I_2(W) = V_{I_1(W)}$.
\end{enumerate}
\end{definition}

\begin{definition}[Word interpretation] Given a Lambek grammar $(\Sigma, \delta, S)$ over a set of basic types $T$ and an interpretation map $I_2 : F(T) \pijl I_2(\delta(\Sigma))$, where $\delta(\Sigma)$ is the relational image of $\Sigma$ under the lexicon, and $I_2(\delta(\Sigma))$ is the image under interpretation (i.e. vector spaces) the word interpretation is a map $I_3$ that respects the following:
\begin{center}
	\begin{tabular}{lcl}
		$I_3(w) \in I_2(W)$ & iff &  $w \ \delta \ W \text{\ and } pol(W) = +$ \\
		$I_3(w) \in I_2(W) \pijl \reals$ & iff & $w \ \delta \ W \text{\ and } pol(W) = -$
	\end{tabular}
\end{center}
That is, words with a positive type are translated as vectors, while words with a negative type are translated as linear maps.
\end{definition}

As an example, if we define the associated vector space of the type $np$ to be $U$ and $n$ to be $P(U)$, then the interpretation of a noun like \quotes{student} will be a constant $I_3(\text{\quotes{student}}) \in V_{P(U)}$, whereas a word like \quotes{all} that is typed $np \s n$ will be a linear map $I_3(\text{\quotes{all}}) \in V_{P(U)} \tensor V_{P(U)} \pijl \reals$.

We proceed to define how we interpret proof terms. The intuitive idea is that a proof term is translated into a linear map that will subsequently be applied to the word interpretations of its antecedents. Though the proof system builds up terms with potentially unbound variables, we require for grammaticality (see above) that the conclusion formula be in focus; this means that the only unbound variables in the proof term are those of the antecedent formula, which will be substituted by word interpretations.

\begin{definition}[Proof term interpretation] Given a proof in the focussed sequent calculus for \textbf{NL}, there is a proof term that encodes the proof. We define the interpretation of a proof by giving the translation of proof terms into linear maps:

\begin{center}
	\begin{tabular}{lcl}
		$Ax(A,x)$ & $\overset{I_4}{\Longrightarrow}$ & $x \in I_3(A)$ \\
		$CoAx(A,\alpha)$ & $\overset{I_4}{\Longrightarrow}$ & $\alpha \in I_3(A)$ \\
		$\leftharpoondown(M,x,A)$ & $\overset{I_4}{\Longrightarrow}$ & $| \{ x \in I_3(A) | I_4(M) \neq 0 \} \ket$ \\
		$\leftharpoonup(M,x)$ & $\overset{I_4}{\Longrightarrow}$ & $x(I_4(M))$ \\
		$\rightharpoonup(M,\alpha)$ & $\overset{I_4}{\Longrightarrow}$ & $\alpha(I_4(M))$ \\
		$\rightharpoondown(M,\alpha)$ & $\overset{I_4}{\Longrightarrow}$ & $\alpha \mapsto I_4(M)$ \\
		$\s L(M,N)$ & $\overset{I_4}{\Longrightarrow}$ & $I_4(M) \tensor I_4(N)$ \\
		$\s R(M,x,\alpha,\beta)$ & $\overset{I_4}{\Longrightarrow}$ & $I_4(M)[\beta \pijl \alpha \tensor x]$ \\
		$\tensor L(M,x,y,z)$ & $\overset{I_4}{\Longrightarrow}$ & $I_4(M)[z \pijl x \tensor y]$ \\
		$\tensor R(M,N)$ & $\overset{I_4}{\Longrightarrow}$ & $I_4(M) \tensor I_4(N)$ \\
		$\bs L(M,N)$ & $\overset{I_4}{\Longrightarrow}$ & $I_4(M) \tensor I_4(N)$ \\
		$\bs R(M,x,\alpha,\beta)$ & $\overset{I_4}{\Longrightarrow}$ & $I_4(M)[\beta \pijl x \tensor \alpha]$
	\end{tabular}
\end{center}
\end{definition}

Finally, as the translation is a continuation-passing-style translation, we will end up with a map that need an evaluation context before finishing computation. So, given that a proof gives a linear map, we apply it to the identity map, and we instantiate the unbound variables with the relevant \emph{word interpretations}.


\subsection{Deriving quantifier scope ambiguity}

Quantifier scope ambiguity as exemplified by the phrase \quotes{Every student likes some teacher} is already shown to be obtainable using the two-step translation process in \cite{bernardi2010continuation} in a Lambek-Grishin grammar, and in a Lambek grammar \cite{moortgat2011proof}. Here, we alter the latter example given to translate into the vector space model as employed by \cite{hedges2016generalised,sadrzadeh2016quantifier} to show that both readings (narrow/wide and wide/narrow) can be obtained and give exactly the kind of meaning we would expect from a vector space model. This means that we can obtain the intended meaning in a \emph{derivational} way. What is more, given that we have both a grammar available and we have learned concrete vectors, the process can potentially be fully automated.
	Each word has to be associated with a syntactic type, and we have to give a word interpretation mapping the words to a vector or linear map. We assume a set of basic types $\{ np, n, s\}$ where $s$ is the distinguished goal type. Polarity assignment is handled by stipulating that $np$ and $n$ are positive, and $s$ is negative. Basic types $np$ and $n$ are interpreted as $U$ and $P(U)$, respectively, and $s$ gets translated to $\reals$. The syntactic types and the word interpretation is given by the following table: \\
\ \\
{\def\arraystretch{1.4}\tabcolsep=7pt
\begin{tabular}{c|c|c}
	$w$ & $\delta(w)$ & $\lceil w \rceil$ \\ 
	\hline
	every	& $np \s n$ & $\epsilon_{\powersetvectorspace} \comp (\meaninginterpretation{\text{all}} \tensor \mu_{\powersetvectorspace}) \comp (\delta_{\powersetvectorspace} \tensor id_{\powersetvectorspace}) \comp \sigma$ \\ 
	student & $n$ & $\means{student}$ \\ 
	likes	& $(np \bs s) \s np$ & $\vektor{a} \tensor f \tensor \vektor{b} \mapsto f(\meaninginterpretation{(\text{likes}_b)_a})$ \\ 
	some	& $np \s n$ & $\epsilon_{\powersetvectorspace} \comp (\meaninginterpretation{\text{some}} \tensor \mu_{\powersetvectorspace}) \comp (\delta_{\powersetvectorspace} \tensor id_{\powersetvectorspace}) \comp \sigma$ \\ 
	teacher	& $n$ & $\means{teacher}$ \\ 
\end{tabular}
}
\ \\[0.5em]
\noindent As a reminder, we also note the vectorial interpretation of lexical constants in the word interpretation:

\begin{center}
{\def\arraystretch{1.4}\tabcolsep=10pt
\begin{tabular}{rcl}
	$\means{all}(|A\ket)$ & $=$ & $\sum\limits_{A \subseteq B \subseteq U} |B\ket$ \\
	$\means{some}(|A\ket)$ & $=$ & $\sum\limits_{\substack{B \subseteq U \text{\ s.t.} \\ A \cap B \neq \emptyset}} |B\ket$ \\
	$\means{student}$ & $=$ & $|A\ket$ for some $A \subseteq U$ \\
	$\means{teacher}$ & $=$ & $|B\ket$ for some $B \subseteq U$ \\
	$\means{likes}$ & $=$ & $\sum\limits_{ij} c_{ij} (|A_i\ket \tensor 1 \tensor |A_j\ket$ for each $A_x \subseteq U$ \\	
\end{tabular}
}
\end{center}

\noindent The two proofs that we get from the focussed sequent calculus are displayed in Figures \ref{fig:widenarrow} and \ref{fig:narrowwide} (without labelling).

If we take the proof term for the first proof and translate this into a vectorial map we get
$$ (1a) \ \alpha \mapsto x \ ( \ | \{ a \in U \ | \ u \ ( \ | \{b \ | \ z \ (a \tensor \alpha \tensor b)) \neq 0 \} \ket \tensor w) \neq 0 \} \ket \tensor y) $$
For the second proof term, we get a slightly different map:
$$ (2a) \ \alpha \mapsto u \ ( \ | \{ a \in U \ | \ x \ ( \ | \{b \ | \ z \ (b \tensor \alpha \tensor a)) \neq 0 \} \ket \tensor y) \neq 0 \} \ket \tensor w) $$

\noindent The unfolded maps are quite intimidating so the complete computation is taken up in the appendix. Here we just note that the two maps reduce to the readings shown below:
{\small
$$ (1b) \ \braket{\means{student}}{| \means{student} \cap \{ a \in U \ | \sum\limits_{\substack{B \subseteq U \text{\ s.t.} \\ \means{teacher} \cap B \neq \emptyset}} \braket{B}{\means{teacher} \cap C} \} \ket}$$
$$\text{where } C = \{ b \in U | \meaninginterpretation{(\text{likes}_b)_a} \neq 0 \}$$
}
{\small
$$ (2b) \ \sum\limits_{\substack{B \subseteq U \text{\ s.t.} \\ \means{teacher} \cap B \neq \emptyset}} \braket{B}{\means{teacher} \cap \{ a \in U \ | \ \braket{\means{student}}{\means{student} \cap D} \neq 0 \}}$$
$$\text{where } D = \{b \in U \ | \ \meaninginterpretation{(\text{likes}_a)_b} \neq 0 \}$$
}
\ \\
\noindent We can see that these interpretations will give different results depending on the instantiation of the vectors. In fact, these interpretations correspond to the result of \cite{sadrzadeh2016quantifier}. This effectively shows that quantifier scope ambiguity can be achieved in vector space models by the use of appropriate proof-theoretic notions.

\section{Concluding Remarks}

In this paper, we elaborated on quantifier scope ambiguity in compositional distributional models of meaning. In particular, the approach of \cite{moortgat2011proof} using a continuation-passing-style translation for a polarised and focussed proof system for the Lambek calculus was combined with the approach to generalised quantifiers of \cite{hedges2016generalised}. The result is a fully derivational and provides a fully worked out compositional way to obtain ambiguous meaning for phrases like \quotes{Every student likes some teacher}, thereby resolving the issue of manually assigning appropriate meaning vectors to such phrases. 

Although we illustrate with examples of two generalised quantifiers in a sentence, the approach works for a single quantifier, and since the applied strategy exploits the combinatorial choices of the proof system (focus on the first quantifier and then on the second one, or vice versa) we expect the approach to generalise to more quantifiers, though the possibility of overgeneration needs to be investigated.

As for experimental validation, since the writing of this paper, it has been recognised that using a powerset construction in vector spaces, to be able to make use of bialgebras, may not be very feasible in practical models: having a powerset as a basis may lead to an exponential blowup in vector space size, and could potentially give sparsity issues. One approach to deal with this could be to use fuzzy quantification \cite{zadeh1983}, which has already been explored by \cite{dostal2016many}.

Another interesting avenue is to work out how several phenomena involving the copying of linguistic material can be analysed in a compositional distributional model. Coordination and pronoun relativisation have been given an account using Frobenius algebras over vector spaces \cite{kartsaklis2016coordination,sadrzadeh2013frobenius}, where the Frobenius operations allow one to express element wise multiplication on arbitrary tensors. In future work we hope to analyse ellipsis, a phenomenon for which it can be argued that copying has to be part of the syntactic process. Rules of controlled copying then can be interpreted using the Frobenius or bialgebra operations. A first step has already been taken by \cite{kartsaklisverb}, and we wish to approach the problem from the typelogical perspective.

\section{Acknowledgments}

I am grateful for a range of insightful discussions with Michael Moortgat on the focussing for the Lambek calculus, and linear non-linearity. Furthermore, I would like to thank Mehrnoosh Sadrzadeh and Dimitri Kartsaklis for the many short and long discussions on Frobenius algebras and bialgebras. Finally, I am grateful for technical comments from Paulo Oliva, and the anonymous reviewers of JLM. I was supported by a Queen Mary Principal Studentship during the writing of this paper. All remaining errors are my own.

\bibliographystyle{plain}
\bibliography{\references}

\appendix

\begin{landscape}
$ \mathbf{(1a)} \ \alpha \mapsto x \ ( \ | \{ a \in U \ | \ u \ ( \ | \{b \in U \ | \ z \ (a \tensor \alpha \tensor b)) \neq 0 \} \ket \tensor w) \neq 0 \} \ket \tensor y) $ \\[1em]

Which, after lexical insertion gives

\begin{center}
	$ \ \alpha \mapsto \lceil\text{every}\rceil \ ( \ | \{ a \in U \ | \ \lceil\text{some}\rceil \ ( \ | \{b \in U \ | \ \lceil\text{likes}\rceil \ (a \tensor \alpha \tensor b)) \neq 0 \} \ket \tensor \lceil\text{teacher}\rceil) \neq 0 \} \ket \tensor \lceil\text{student}\rceil) $
\end{center}

Unfolding the definition and inserting the identity map gives \\[1.5em]
{\small
$ \epsilon_{\powersetvectorspace} \ \comp \ \big(\meaninginterpretation{\text{all}} \tensor \mu_{\powersetvectorspace}\big) \ \comp \ \big(\delta_{\powersetvectorspace} \tensor id_{\powersetvectorspace}\big)$\\[0.5em]
$\qquad \qquad \comp \ \sigma \Big( \ \big| \big\{ a \in U \ | \ \epsilon_{\powersetvectorspace} \comp \big(\meaninginterpretation{\text{some}} \tensor \mu_{\powersetvectorspace}\big) \ \comp \ \big(\delta_{\powersetvectorspace} \tensor id_{\powersetvectorspace}\big) \comp \ \sigma \big( \ | \{b \in U \ | \ \meaninginterpretation{(\text{likes}_b)_a} \neq 0 \} \ket \tensor \means{teacher}\big) \neq 0 \big\} \big\ket \tensor |\means{student}\ket\Big) $ \\[1em]
\ \\
$= \epsilon_{\powersetvectorspace} \ \comp \ \big(\meaninginterpretation{\text{all}} \tensor \mu_{\powersetvectorspace}\big) \ \comp \ \big(\delta_{\powersetvectorspace} \tensor id_{\powersetvectorspace}\big) \ \comp \ \sigma \Big(\ \big| \big\{ a \in U \ | \sum\limits_{\substack{B \subseteq U \text{\ s.t.} \\ \means{teacher} \cap B \neq \emptyset}} \braket{B}{\means{teacher} \cap \{ b \in U | \meaninginterpretation{(\text{likes}_b)_a} \neq 0 \} } \big\}\big\ket \tensor \means{student}\Big) $\\[1.5em]

$= \sum\limits_{\means{student} \subseteq C \subseteq U} \braket{C}{| \means{student} \cap \{ a \in U \ | \sum\limits_{\substack{B \subseteq U \text{\ s.t.} \\ \means{teacher} \cap B \neq \emptyset}} \braket{B}{\means{teacher} \cap \{ b \in U | \meaninginterpretation{(\text{likes}_b)_a} \neq 0 \} } \} \ket}$\\

}
\label{fig:reading1}
\end{landscape}

\appendix

\begin{landscape}
	
$ \mathbf{(2a)} \ \alpha \mapsto u \ ( \ | \{ a \in U \ | \ x \ ( \ | \{b \in U \ | \ z \ (b \tensor \alpha \tensor a)) \neq 0 \} \ket \tensor y) \neq 0 \} \ket \tensor w) $ \\[1em]

Which, after lexical insertion gives

\begin{center}
$ \ \alpha \mapsto \lceil\text{some}\rceil \ ( \ | \{ a \in U \ | \ \lceil\text{every}\rceil \ ( \ | \{b \in U \ | \ \lceil\text{likes}\rceil \ (b \tensor \alpha \tensor a)) \neq 0 \} \ket \tensor \lceil\text{student}\rceil) \neq 0 \} \ket \tensor \lceil\text{teacher}\rceil) $
\end{center}

Unfolding the definition and inserting the identity map gives \\[1.5em]
{\small
$ \epsilon_{\powersetvectorspace} \ \comp \ \big(\meaninginterpretation{\text{some}} \tensor \mu_{\powersetvectorspace}\big) \ \comp \ \big(\delta_{\powersetvectorspace} \tensor id_{\powersetvectorspace}\big)$\\[0.5em]
$\comp \ \sigma \Big( \ \Big| \Big\{ a \in U \ | \ \epsilon_{\powersetvectorspace} \ \comp \ \big(\meaninginterpretation{\text{all}} \tensor \mu_{\powersetvectorspace}\big) \ \comp \ \big(\delta_{\powersetvectorspace} \tensor id_{\powersetvectorspace}\big) \ \comp \ \sigma \big( \ | \{b \in U \ | \ \meaninginterpretation{(\text{likes}_a)_b} \neq 0 \} \ket \tensor \means{student} \big) \neq 0 \Big\} \Big\ket \tensor \means{teacher}\Big) $ \\[1em]
\ \\
$= \epsilon_{\powersetvectorspace} \comp (\meaninginterpretation{\text{some}} \tensor \mu_{\powersetvectorspace}) \comp (\delta_{\powersetvectorspace} \tensor id_{\powersetvectorspace}) \ \comp \ \sigma \Big(\ \Big| \Big\{ a \in U \ | \ \sum\limits_{\means{student} \subseteq B \subseteq U} \braket{B}{\means{student} \cap \{b \in U \ | \ \meaninginterpretation{(\text{likes}_a)_b} \neq 0 \}} \neq 0 \Big\} \Big\ket \tensor \means{teacher} \Big)$\\[1.5em]

$= \sum\limits_{\substack{C \subseteq U \text{\ s.t.} \\ \means{teacher} \cap C \neq \emptyset}} \braket{C}{\means{teacher} \ \cap \ \{ a \in U \ | \ \sum\limits_{\means{student} \subseteq B \subseteq U} \braket{B}{\means{student} \ \cap \ \{b \in U \ | \ \meaninginterpretation{(\text{likes}_a)_b} \neq 0 \}} \neq 0 \}}$\\

}
\label{fig:reading2}
\end{landscape}

\appendix

\begin{landscape}
\begin{figure}
$$\infer[\rightharpoondown]{\seq{(\underset{\red{every}}{x : np \slash n} \bullet \underset{\red{student}}{y : n}) \bullet (\underset{\red{likes}}{z : (np \bs s) \slash np} \bullet (\underset{\red{some}}{u : np \slash n} \bullet \underset{\red{teacher}}{w : n)})}{\focus{s}}}{
\infer[\leftharpoonup]{
\seq{(x : np \slash n \bullet y : n) \bullet (z : (np \bs s) \slash np \bullet (u : np \slash n \bullet w : n))}{\alpha : s}}
{
\infer[\slash L]{\seq{(\focus{np \slash n} \bullet y : n) \bullet (z : (np \bs s) \slash np \bullet (u : np \slash n \bullet w : n))}{\alpha : s}}{
\infer[\leftharpoondown]{\seq{\focus{np} \bullet (z : (np \bs s) \slash np \bullet (u : np \slash n \bullet w : n))}{\alpha : s}}{
\infer[\leftharpoonup]{\seq{a : np \bullet (z : (np \bs s) \slash np \bullet (u : np \slash n \bullet w : n))}{\alpha : s}}{
\infer[\slash L]{\seq{a : np \bullet (z : (np \bs s) \slash np \bullet (\focus{np \slash n} \bullet w : n))}{\alpha : s}}{\infer[\leftharpoondown]{\seq{a : np \bullet (z : (np \bs s) \slash np \bullet \focus{np})}{\alpha : s}}{ \infer[\leftharpoonup]{\seq{a : np \bullet (z : (np \bs s) \slash np \bullet b : np)}{\alpha : s}}{\infer[\slash L]{\seq{a : np \bullet (\focus{(np \bs s) \slash np} \bullet b : np)}{\alpha : s}}{\infer[\bs L]{\seq{a : np \bullet \focus{np \bs s}}{\alpha : s}}{\infer[Ax]{\seq{a : np}{\focus{np}}}{} & \infer[CoAx]{\seq{\focus{s}}{\alpha : s}}{}} & \infer[Ax]{\seq{b : np}{\focus{np}}}{}}}} & \infer[Ax]{\seq{w : n}{\focus{n}}}{\red{teacher}}}}} & \infer[Ax]{\seq{y : n}{\focus{n}}}{\red{student}}}}}$$
\caption{Proof for wide over narrow scope}
\label{fig:widenarrow}
\end{figure}
\end{landscape}

\appendix

\begin{landscape}
\begin{figure}
$$\infer[\rightharpoondown]{\seq{(\underset{\red{every}}{x : np \slash n} \bullet \underset{\red{student}}{y : n}) \bullet (\underset{\red{likes}}{z : (np \bs s) \slash np} \bullet (\underset{\red{some}}{u : np \slash n} \bullet \underset{\red{teacher}}{w : n)})}{\focus{s}}}{\infer[\leftharpoonup]{\seq{(x : np \slash n \bullet y : n) \bullet (z : (np \bs s) \slash np \bullet (u : np \slash n \bullet w : n))}{\alpha : s}}{
\infer[\slash L]{\seq{(x : np \slash n \bullet y : n) \bullet (z : (np \bs s) \slash np \bullet (\focus{np \slash n} \bullet w : n))}{\alpha : s}}{
\infer[\leftharpoondown]{\seq{(x : np \slash n \bullet y : n) \bullet (z : (np \bs s) \slash np \bullet \focus{np})}{\alpha : s}}{\infer[\leftharpoonup]{\seq{(x : np \slash n \bullet y : n) \bullet (z : (np \bs s) \slash np \bullet a : np)}{\alpha : s}}{
\infer[\slash L]{\seq{(\focus{np \slash n} \bullet y : n) \bullet (z : (np \bs s) \slash np \bullet a : np)}{\alpha : s}}{
\infer[\leftharpoondown]{\seq{\focus{np} \bullet (z : (np \bs s) \slash np \bullet a : np)}{\alpha : s}}{
\infer[\leftharpoonup]{\seq{b : np \bullet (z : (np \bs s) \slash np \bullet a : np)}{\alpha : s}}{\infer[\slash L]{\seq{b : np \bullet (\focus{(np \bs s) \slash np} \bullet a : np)}{\alpha : s}}{
\infer[\bs L]{\seq{b : np \bullet \focus{np \bs s}}{\alpha : s}}{\infer[Ax]{\seq{b : np}{\focus{np}}}{} & \infer[CoAx]{\seq{\focus{s}}{\alpha : s}}{}} & \infer[Ax]{\seq{a : np}{\focus{np}}}{}}}} & \infer[Ax]{\seq{y : n}{\focus{n}}}{\red{student}}}}} & \infer[Ax]{\seq{w : n}{\focus{n}}}{\red{teacher}}}}}$$
\caption{A proof for narrow over wide scope}
\label{fig:narrowwide}
\end{figure}
\end{landscape}

\end{document}